# Safety-Aware Human-Lead Vehicle Platooning
# by Proactively Reacting to Uncertain Human Behaving

**Jia Hu, Ph.D.**
ZhongTe Distinguished Chair in Cooperative Automation, Professor
Key Laboratory of Road and Traffic Engineering of the Ministry of Education
Tongji University, Shanghai, China, 201804
the State Key Laboratory of Advanced Design and Manufacturing Technology for Vehicle
Tel: +86-13588159138; Email: hujia@tongji.edu.cn

**Shuhan Wang**
Key Laboratory of Road and Traffic Engineering of the Ministry of Education
Tongji University, Shanghai, China, 201804
Tel: +86-18851658012; Email: shu_hans@tongji.edu.cn

**Yiming Zhang**
Key Laboratory of Road and Traffic Engineering of the Ministry of Education
Tongji University, Shanghai, China, 201804
Tel: +86-18916388013; Email: z-emin@foxmail.com

**Haoran Wang, Corresponding Author**
Associate Researcher
Key Laboratory of Road and Traffic Engineering of the Ministry of Education, Tongji University, 4800 Cao'an Road, Shanghai, P.R.China
State Key Laboratory of Advanced Design and Manufacturing for Vehicle Body, Hunan University, Changsha, 410082, China
Email: wang_haoran@tongji.edu.cn

**Zhilong Liu**
Senior Engineer
Dazhuo Intelligent Technology Company, Shanghai, P.R.China
Email: zhilongliu@mychery.com

**Guangzhi Cao**
Chief Technology Officer
ZDrive.ai, Shanghai, P.R.China
Email: gary@mychery.com

**ABSTRACT**

Human-Lead Cooperative Adaptive Cruise Control (HL-CACC) is regarded as a promising vehicle platooning technology in real-world implementation. By utilizing a Human-driven Vehicle (HV) as the platoon leader, HL-CACC reduces the cost and enhances the reliability of perception and decision-making. However, state-of-the-art HL-CACC technology still has a great limitation on driving safety due to the lack of considering the leading human driver's uncertain behavior. In this study, a HL-CACC controller is designed based on Stochastic Model Predictive Control (SMPC). It is enabled to predict the driving intention of the leading Connected Human-Driven Vehicle (CHV). The proposed controller has the following features: i) enhanced perceived safety in oscillating traffic; ii) guaranteed safety against hard brakes; iii) computational efficiency for real-time implementation. The proposed controller is evaluated on a PreScan&Simulink simulation platform. Real vehicle trajectory data is collected for the calibration of the simulation. Results reveal that the proposed controller: i) improves perceived safety by 19.17% in oscillating traffic; ii) enhances actual safety by 7.76% against hard brakes; iii) is confirmed with string stability. The computation time is approximately 3.2 milliseconds when running on a laptop equipped with an Intel i5-13500H CPU. This indicates the proposed controller is ready for real-time implementation.

# 1 INTRODUCTION

Cooperative Adaptive Cruise Control (CACC) forms Connected Automated Vehicles (CAVs) into a platoon with much smaller following headway between adjacent CAVs (Zhou et al., 2017). Hence, CACC is proven to enhance both roadway capacity (Ge and Orosz, 2018) and fuel efficiency (Vahidi and Eskandarian, 2003). However, CACC is hard to be applied in commercial operations, because there are no proven autonomous driving technologies capable of safely leading a CACC platoon on open roads (Wang et al., 2022). A promising alternative, Human-Lead CACC (HL-CACC), has emerged to address this problem. In the HL-CACC system, the leader of a platoon is a Connected Human-Driven Vehicle (CHV). By taking advantage of human expertise in perception and decision-making, HL-CACC facilitates the practical application of CACC in the real world. Currently, nearly all OEMs and Tier 1 suppliers adopt the HL-CACC mechanism in the commercial operation of CACC, including our collaboration with SAIC at Shanghai Yangshan Port (Zhang et al., 2022).

The HL-CACC still has safety risks due to the disturbance of the leading CHV's uncertain behavior (Deng et al., 2023; Hajdu et al., 2020). The leading CHV's reactions to the environment may be either overreacting or delayed responding (Du et al., 2022). The overreaction amplifies traffic speed oscillations (Li et al., 2023a; Saifuzzaman et al., 2017), bringing perceived risks (Tian et al., 2021) and reducing string instability (Bouadi et al., 2022; Li et al., 2023b). Conversely, the delayed response may fail to timely react to the environment, resulting in hard brakes and even collisions (Ding et al., 2019; Kiefer et al., 2005). Hence, the uncertain behaving of human drivers is crucial to be addressed before the implementation of HL-CACC.

A few studies have dealt with the uncertainty of human behaving in a HL-CACC platoon. Bichi (Bichi et al., 2010) modeled the one-step prediction of the leading CHV's acceleration based on the discretized Markov Chain (MC). This method, however, relies on a preset distribution of the driver's acceleration, making the prediction inaccurate in practice. To match up with human's actual intentions, car following models are extensively used. Gong and Du (Gong and Du, 2018) used Newell car-following models to detect the movement of CHVs. Zhao (Zhao et al., 2018) employed an Optimal Velocity Model (OVM) to describe the behavior of CHVs in a cooperative eco-driving problem. However, these traffic-flow dynamics models are deterministic, thereby unable to depict the inherent stochasticity of human behaving (Ngoduy et al., 2019). Artificial intelligence methods have been focused in recent years. Yang and Chu (Yang et al., 2023) employ LSTM networks to predict the acceleration of the leading CHV. Ozkan (Ozkan and Ma, 2022) proposed a distributed stochastic control based CACC method, utilizing a stochastic inverse reinforcement learning approach for the prediction of the leading CHV. However, these artificial intelligence methods can only output the prediction results which are most likely to happen. A lot of risky prediction results yet with low possibility are overlooked (He and Lv, 2023). These long-tail prediction results are quite crucial to be considered, to enable the safe implementation of HL-CACC.

Therefore, this paper proposed a Stochastic Driver Model based Human-Lead CACC (SDHL-CACC) controller. It formulates a scenario-based Stochastic Model Predictive Control (SMPC) problem (Bernardini and Bemporad, 2009), to consider all uncertain behaving of the leading CHV. The proposed SDHL-CACC controller has the following features:
- Enhanced perceived safety in oscillating traffic;
- Guaranteed safety against hard brakes;
- Computational efficiency for real-time implementation.

The rest of this paper is organized as follows. Section II describes the highlights of this research; section III presents the proposed controller; section IV evaluates the proposed controller and analyzes the results of simulation tests. Section V provides further discussion and outlines further directions.

## 2 HIGHLIGHTS

To specifically bear the aforementioned features, the proposed SDHL-CACC controller takes the following methodological contributions:

- **Look before the leap**

All possible actions of the leading CHV are considered by the proposed SDHL-CACC controller. It is realized by a scenario-based SMPC approach. In this framework, the generated scenarios are capable of covering all possible actions of the leading CHV. Planning is conducted based on the weighted evaluation of all scenarios. Hence, the proposed SDHL-CACC controller is enabled to fully look before the leap.

- **Contingency plan for long-tail risks**

To guarantee actual safety, the proposed SDHL-CACC controller is enabled to contingency plan for long-tail risks. A Conditional Value-at-Risk (CVaR) method is adopted to measure the severity of all possible risks. In this way, severe risks with small probability are still prioritized in the planning. Hence, the proposed controller greatly reduces dangerous actions of a platoon, like hard brakes.

- **Convex formulation**

The proposed SDHL-CACC controller is transformed into a standard quadratic programming problem with linear constraints. It ensures the computation efficiency of the proposed controller.

## 3 PROBLEM FORMULATION

HL-CACC is illustrated in **Fig. 1**. In an HL-CACC system, the platoon leader is a CHV and the followers are CAVs. The leading CHV can only provide its current state and past maneuver information. The leading CHV's future motions are uncertain for the followers. It may cause disturbance on its following CAVs, such as stochastic speed fluctuation and drastic speed reduction to avoid collisions.

The Predecessor-Leader Following (PLF) communication topology is adopted (Dey et al., 2016; Peters et al., 2014). Under this type of topology, each following CAV communicates with both its predecessor and the leading CHV. The control objective of an HL-CACC system is to minimize all CAVs' tracking errors with respect to both the predecessor and the leader. The constant distance following strategy is employed in this research, where each CAV maintains a fixed distance with respect to its preceding vehicle. This strategy is particularly effective for highway driving, as it ensures a small gap even at high speeds. It would enhance communication quality, providing reliable data exchange in the platoon.

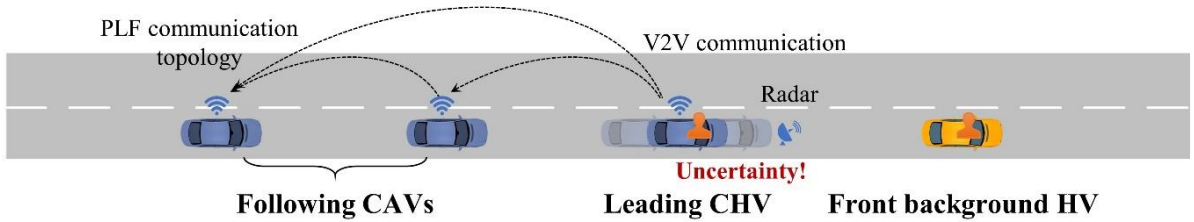

**Fig. 1.** The scenario of interest: a human-lead platoon with a background HV in the front

### 3.1 Control Architecture

The system structure of the proposed SDHL-CACC controller is illustrated in **Fig. 2**.

- **Stochastic driver model**: This model is utilized to predict the uncertain behaving of the leading CHV according to the real-time traffic environment. The stochastic driver model is calibrated by real vehicle trajectory data. The prediction results are provided to the platoon controller in the form of a scenario tree.
- **Scenario-based SMPC controller**: This controller calculates the optimal action of each CAV follower based on the scenario tree. It outputs the expected control commands to CAV's local control for execution.

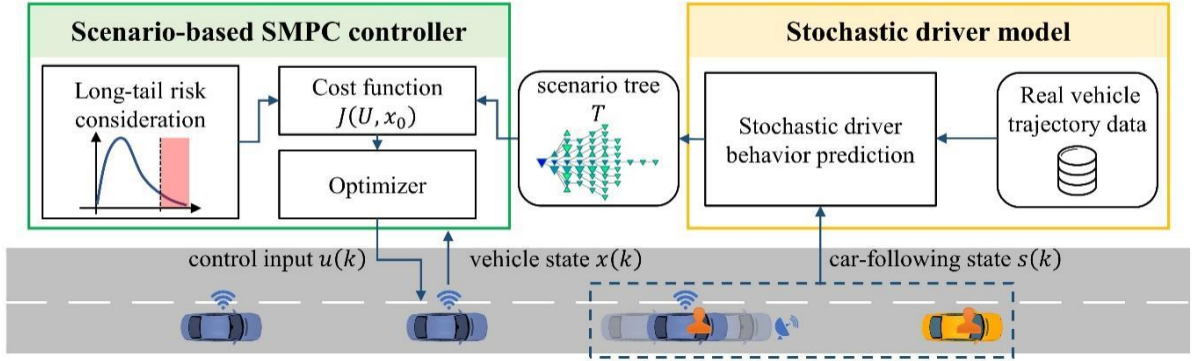

**Fig. 2.** Structure of the proposed SDHL-CACC controller

### 3.2 System Dynamics

Based on the PLF communication topology, each CAV follows its predecessor and the platoon leader. The state vector and control vector of a CAV follower are defined in Definition 1. The desired state of the ego CAV is presented in Definition 2.

***Definition 1*** (State and control vector): The ego CAV's system state vector $x$ and the control vector $u$ are defined as follows:

$$x = \begin{bmatrix} h_L^* - h_L \\ \Delta v_L \\ h_P^* - h_P \\ \Delta v_P \\ a \end{bmatrix} \quad (1)$$

$$u = u_a \quad (2)$$

with

$$\Delta v_L = v_L - v \quad (3)$$
$$\Delta v_P = v_P - v \quad (4)$$

where $h_L^*$ is the desired distance headway with the leading CHV, $h_L$ is the actual distance headway with the leading CHV, $h_P^*$ is the desired distance headway with the proceeding vehicle, $h_P$ is the actual distance headway with the proceeding vehicle, $v_L$ is the velocity of the leading CHV, $v_P$ is the velocity of the proceeding vehicle, $v$ is the velocity of the ego vehicle, $a$ is the acceleration of the ego vehicle, $u_a$ is the desired acceleration of the ego vehicle.

***Definition 2*** (Desired state): The ego CAV's desired state $x_{ref}$ is defined as follows:

$$x_{ref} = (0,0,0,0,0)^T \quad (5)$$

The dynamics of the following CAVs are modeled by a linear system with an additive stochastic disturbance vector as follows:

$$x(k+1) = A_k x(k) + B_k u(k) + C_k w(k) \qquad (6)$$

with

$$A_k = I_{5*5} + \begin{bmatrix} 0 & -1 & 0 & 0 & 0 \\ 0 & 0 & 0 & 0 & -1 \\ 0 & 0 & 0 & -1 & 0 \\ 0 & 0 & 0 & 0 & -1 \\ 0 & 0 & 0 & 0 & -\dfrac{1}{\tau_a} \end{bmatrix} * \Delta t \qquad (7)$$

$$B_k = \begin{bmatrix} 0 \\ 0 \\ 0 \\ 0 \\ \dfrac{1}{\tau_a} \end{bmatrix} * \Delta t \qquad (8)$$

$$C_k = \begin{bmatrix} 0 & 0 \\ 1 & 0 \\ 0 & 0 \\ 0 & 1 \\ 0 & 0 \end{bmatrix} * \Delta t \qquad (9)$$

$$w = \begin{bmatrix} a_L \\ a_P \end{bmatrix} \qquad (10)$$

where $\tau_a$ is the first-order inertial delay parameter of the ego vehicle's system as actuation delay is taking into consideration (Zhang et al., 2022), $w$ is the additive multi-dimensional disturbance vector. In the stochastic dynamics, the leader acceleration $a_L$ and the predecessor acceleration $a_P$ are assumed to be the stochastic disturbances of the system.

### 3.3 Stochastic Driver Model

A stochastic driver model is used to predict the acceleration of the leading CHV. The model formulation is detailed in the remainder of this section.

To describe the stochasticity of the speed oscillation of the leading CHV, a stochastic car-following model based on Langevin equations is applied in this study. Langevin equations illustrate the stochastic process in physics. They are used to describe the time-varying random acceleration of an HV with respect to its front vehicle (Tian et al., 2016).

In this paper, the stochastic car-following model of the leading CHV is formulated as equation (11). It consists of a deterministic car-following model $\beta[v_{op}(s(k)) - v_L(k)]$ (Bando et al., 1995), and a Langevin stochastic model $\sigma_0 \sqrt{v_L(k)} \Delta W(k)$.

$$a_L(k+1) = \beta[v_{op}(s(k)) - v_L(k)] + \sigma_0 \sqrt{v_L(k)} \Delta W(k) \qquad (11)$$

where $v_L(k)$ is the speed of the platoon leader, $\beta$ is the reaction coefficient.

The deterministic car-following model is adopted to calculate the mean speed of the leading CHV. The only required information from the front background HV is the space headway, which could be easily sensed by the onboard radar on the leading CHV. In this model, $v_{op}(s(k))$ is the optimal speed, defined as:

$$v_{op}(s(k)) = \frac{v_0}{2}\left[\tanh\left(\frac{s(k)}{s_c} - \alpha\right) + \tanh\alpha\right] \tag{12}$$

where $v_0$ is the free flow speed, $s(k)$ is the distance headway between the leading CHV and the front background HV, $s_c$ is the critical headway, $\alpha$ is the dimensionless constant coefficient.

In the Langevin stochastic model, $\Delta W(k)$ follows a Wiener process, which is adopted to describe the random acceleration deviations. Based on an extended Cox-Ingersoll-Ross (CIR) process (Cox et al., 1985), $\sigma_0\sqrt{v_L(k)}$ is introduced as a dissipation factor to reflect the noise strength of acceleration, where $\sigma_0$ is the dissipation coefficient. The randomness in speed change depends on the arbitrary values of the leading CHV's velocity. Higher speed implies more randomness in the acceleration.

At each timestep $k$, the leading CHV's acceleration at the next timestep $a_L(k+1)$ is predicted by the information about the leading CHV and its front vehicle based on equation (11). The continuous range of the leading CHV's future possible acceleration is divided into $m$ multiple discrete intervals which are represented by $m$ several discrete values:

$$A_L(k+1) = \{a_L^1(k+1), a_L^2(k+1), \cdots, a_L^m(k+1)\} \tag{13}$$

with a set of time-varying probability:

$$p(k) = \{p_1(k), p_2(k), \cdots, p_m(k)\} \tag{14}$$

where $p_i(k) = \Pr[a_L(k) = a_i(k)|v_L(k-1), s(k-1)]$ with $\sum_{i=1}^m p_i(k) = 1$ for $k \in Z_+$ and $i = 1, 2, \cdots, m$.

### 3.4 Controller Design

In this section, the SMPC approach based on scenario enumeration and multi-stage stochastic optimization is applied (Bernardini and Bemporad, 2009). The SMPC optimization problem is formulated by a maximum likelihood approach. In this way, an optimization tree is introduced to describe the stochastic evolution of the disturbance in the future.

The scenario tree is generated with the following rules. At every timestep, an optimization tree is built according to the current state of the system. Each node of the tree represents a possible scenario of the system, while the root node represents the deterministic measurements of the state at present. The stochastic driver behavior model is adopted to model the probability distribution of the disturbance in each scenario. Starting from the root node, the optimization tree is expanded in the most likely direction by repeating the following steps:

*Step 1*: A list of possible candidate successor nodes of the latest added node is generated by the driver model.

*Step 2*: The realization probability of reaching each new candidate node from the root node is calculated as:

$$\pi_C = \pi_L \times p_L^C \tag{15}$$

where $\pi_C$ is the realization probability of a new candidate node, $\pi_L$ is the realization probability of the latest added node, which is calculated during the previous cycle, $p_L^C$ is the probability of reaching one new candidate node from the latest added node.

**Step 3**: The candidate node with the maximum realization probability from the root node is added to the tree. Note that all candidate nodes are considered in this step.

**Step 4**: Remove the latest added node in **Step 3** from the candidate list and reserve the others.

The above steps are repeated until the tree reaches the desired number of nodes $n_{max}$.

In the optimal control problem, each node represents a control step and every path from the root node to a leaf node represents a prediction horizon. Therefore, a "multiple-horizon" approach is developed, where a longer path represents a larger disturbance realization probability and is attached with more weight when optimizing.

In the optimization tree, the superscript indicates the order in which the nodes are added and the subscript signifies the step in a predictive horizon. The superscript or the subscript is omitted for simplicity in the subsequent section. Thus, the notation related to the optimization tree is as follows:

- $T_r = T_1$ is the root node.
- $T = \{T_1, T_2, \cdots, T_{n_{max}}\}$ is the set of tree nodes that are indexed progressively as they are added to the tree.
- $L$ is the set of leaf nodes.
- $N$ is the number of steps of the longest horizon.
- $N_{nr} = n_{max} - 1$ is the number of nodes except the root nodes.
- $N_{nl}$ is the number of nodes except the leaf nodes, that is, the number of nodes that are not the end node of the specific horizon.
- $x_i^j$ is the state of each node.
- $u_i^j$ is the input control of each node. Note that the expected control input remains constant across the same step of different horizons, that is, for each $k = 0, 1, \cdots, N-1$ and each $j_1, j_2 = 1, 2, \cdots, n_{max}$, $u_k^{j_1} = u_k^{j_2}$.
- $w_i^j$ is the disturbances of each node.

Based on the optimization tree, the stochastic finite-time optimal control function is defined. The state cost is formulated as a quadratic form. To consider long-tail risks, the Conditional Value-at-Risk (CVaR) is introduced to measure the severity of possible risks. Thus, the cost function of the scenario-based stochastic MPC problem can be formulated as:

$$J(U, x_0) = \sum_{i \in T \setminus T_r} \pi^i (x^i - x_{ref})^T Q (x^i - x_{ref}) + \sum_{j \in T \setminus L} \pi^j u^{jT} R u^j + M \sum_{i \in T \setminus T_r} \pi^i \max(x^i(3) - e_r, 0) \quad (16)$$

where $\pi^i$ is the probability of reaching the i$^{th}$ node from the root node, $x^i$ is the state of the scenario related to the i$^{th}$ node, $x^i(3)$ is the third element of $x^i$, the distance headway error between the ego CAV and its predecessor, $e_r$ is the maximum tolerable one-way error, $M$ is a very large weighting factor, $U = [u_0 \ u_1 \ \cdots \ u_{N-1}]^T$ is the input at each step, $Q$ and $R$ are semipositive definite weighting factor matrices.

The cost function implies that the state vector at the i$^{th}$ node is:
$$x^i = A_k x^{pre(i)} + B_k u^{pre(i)} + C_k w^{pre(i)}, i \in T \setminus T_r \quad (17)$$
where $pre(i)$ represents the predecessor node of the i$^{th}$ node.

The vehicle's desired acceleration should be bounded by vehicle capability and comfort.
$$a_{min} \leq u_i \leq a_{max}, for\ i = 0, 1, \cdots, N-1 \quad (18)$$

**3.5 Solution**

This section details the solution of the proposed controller. The proposed SMPC formulation can be transformed into a condensed form, thus transformed into a standard quadratic programming problem.

Consider a single horizon with a series of $N$ nodes, state of each node can be formulated as:
$$X = \begin{bmatrix} x_1 \\ x_2 \\ \vdots \\ x_N \end{bmatrix} = \mathcal{A}x_0 + \mathcal{B}\begin{bmatrix} u_0 \\ u_1 \\ \vdots \\ u_{N-1} \end{bmatrix} + \mathcal{C}\begin{bmatrix} w_0 \\ w_1 \\ \vdots \\ w_{N-1} \end{bmatrix} = \mathcal{A}x_0 + \mathcal{B}U + \mathcal{C}W \quad (19)$$

where
$$\mathcal{A} = \begin{bmatrix} A_k \\ A_k^2 \\ \vdots \\ A_k^N \end{bmatrix} \quad (20)$$

$$\mathcal{B} = \begin{bmatrix} B_k & & & \\ A_k B_k & B_k & & \\ \vdots & \vdots & \ddots & \\ A_k^{N-1} B_k & A_k^{N-2} B_k & \cdots & B_k \end{bmatrix} \quad (21)$$

$$\mathcal{C} = \begin{bmatrix} C_k & & & \\ A_k C_k & C_k & & \\ \vdots & \vdots & \ddots & \\ A_k^{N-1} C_k & A_k^{N-2} C_k & \cdots & C_k \end{bmatrix} \quad (22)$$

In a scenario tree, every horizon is not longer than $N$ steps. Thus, the state of each node in the optimization tree (except the root node) can be formulated as:
$$\mathcal{X} = \begin{bmatrix} x^1 \\ x^2 \\ \vdots \\ x^{N_{nr}} \end{bmatrix} = \bar{\mathcal{A}}x_0 + \bar{\mathcal{B}}\begin{bmatrix} u_0 \\ u_1 \\ \vdots \\ u_{N-1} \end{bmatrix} + \bar{\mathcal{C}}\begin{bmatrix} w^1 \\ w^2 \\ \vdots \\ w^{N_{nl}} \end{bmatrix} = \bar{\mathcal{A}}x_0 + \bar{\mathcal{B}}U + \bar{\mathcal{C}}\mathcal{W} \quad (23)$$

where $\bar{\mathcal{A}}, \bar{\mathcal{B}}, \bar{\mathcal{C}}$ are transformed matrices $\mathcal{A}, \mathcal{B}, \mathcal{C}$ according to the scenario tree, $\mathcal{W}$ is the set of disturbances of all nodes except the leaf nodes.

The violence of safety of each node in the optimization tree (except the root node) can be formulated as:
$$\mathcal{Y} = \begin{bmatrix} x^1(3) - e \\ x^2(3) - e \\ \vdots \\ x^{N_{nr}}(3) - e \end{bmatrix} = \begin{bmatrix} x^1(3) \\ x^2(3) \\ \vdots \\ x^{N_{nr}}(3) \end{bmatrix} - e \times I_{N_{nr} \times 1} = \overline{\mathcal{A}_r}x_0 + \overline{\mathcal{B}_r}U + \overline{\mathcal{C}_r}\mathcal{W} \quad (24)$$

where $\overline{\mathcal{A}_r}, \overline{\mathcal{B}_r}, \overline{\mathcal{C}_r}$ are the set of the third element of the state vector of each node in $\bar{\mathcal{A}}, \bar{\mathcal{B}}, \bar{\mathcal{C}}$.

Thus, the cost function (equation (16)) can be represented in the form of matrices:
$$J(U, x_0) = \mathcal{X}^T \bar{Q} \mathcal{X} + U^T \bar{R} U + \bar{M}^T \max(\mathcal{Y}, 0) \quad (25)$$

where $\bar{Q}, \bar{R}, \bar{M}$ are the integrated weighting factor matrices multiplied with the probability of each node.

After introducing the decision variable $Z = \begin{bmatrix} z_1 \\ z_2 \\ \vdots \\ z_{N_{nr}} \end{bmatrix}$, the cost function (equation (25)) is transformed into a convex optimization problem:

$$J(U, Z, x_0) = \mathcal{X}^T \bar{Q} \mathcal{X} + U^T \bar{R} U + \bar{M}^T Z \tag{26}$$

$$\text{s.t.} \begin{cases} a_{min} \leq u^i \leq a_{max}, \text{for } i = 0, 1, \cdots, N-1 \\ Z \geq \mathcal{Y} \\ Z \geq O_{N_{nr} \times 1} \end{cases} \tag{27}$$

Thus, the former optimization problem is transformed as:

$$\min_{\mathcal{U}} J(\mathcal{U}, x_0) = \mathcal{X}^T \bar{Q} \mathcal{X} + U^T \bar{R} U + \bar{M}^T Z = \frac{1}{2} \mathcal{U}^T H \mathcal{U} + f^T \mathcal{U} + D \tag{28}$$

$$\text{s.t.} \begin{cases} P\mathcal{U} \leq G \\ \begin{bmatrix} a_{min} \times I_{N \times 1} \\ O_{N_{nr} \times 1} \end{bmatrix} \leq \mathcal{U} \leq \begin{bmatrix} a_{max} \times I_{N \times 1} \\ z_{max} \times I_{N_{nr} \times 1} \end{bmatrix} \end{cases} \tag{29}$$

where in the cost function, $\mathcal{U} = \begin{bmatrix} U \\ Z \end{bmatrix}$ is the decision variable, $H = \begin{bmatrix} 2(\bar{\mathcal{B}}^T \bar{Q} \bar{\mathcal{B}} + \bar{R}) & O_{N \times N_{nr}} \\ O_{N_{nr} \times N} & O_{N_{nr} \times N_{nr}} \end{bmatrix}$ is a semipositive definite matrix, $f = \begin{bmatrix} 2(\bar{\mathcal{B}}^T \bar{Q} \bar{\mathcal{A}} x_0 + \bar{\mathcal{B}}^T \bar{Q} \bar{\mathcal{C}} \mathcal{W}) \\ \bar{M} \end{bmatrix}$, $D = (\bar{\mathcal{A}} x_0 + \bar{\mathcal{C}} \mathcal{W})^T \bar{Q} (\bar{\mathcal{A}} x_0 + \bar{\mathcal{C}} \mathcal{W})$ is a constant number, in the constraints, $P = [\overline{\mathcal{B}_r} \quad -E_{N_{nr} \times N_{nr}}]$, $G = e \times I_{N_{nr} \times 1} - \overline{\mathcal{A}_r} x_0 - \overline{\mathcal{C}_r} \mathcal{W}$, $z_{max}$ is an adequate positive number for the actual solution of $Z$ is usually equal or approximately equal to zero.

The SMPC optimization process is formulated as a standard quadratic programming problem with linear constraints. A MATLAB pre-installed QP solver, which uses the trust-region-reflective algorithm (Coleman and Li, 1996), is utilized to calculate the optimal control action in this paper.

## 4 EVALUATION

The performance of the proposed SDHL-CACC controller is evaluated from the following aspects: i) human-lead platooning function validation; ii) perceived safety evaluation in oscillating traffic; iii) actual safety evaluation against hard brakes; iv) platoon string stability validation. The simulation platform is PreScan and Matlab/Simulink coupled. PreScan is a physics-based simulation platform that provides efficient math models of vehicle dynamics behavior.

### 4.1 Experimental Design

*4.1.1 Test Scenarios*

The test scenario is illustrated in **Fig. 3**. In this scenario, the human lead platoon is traveling. It consists of three vehicles: one leading CHV and two following CAVs. The platoon follows a front background HV.

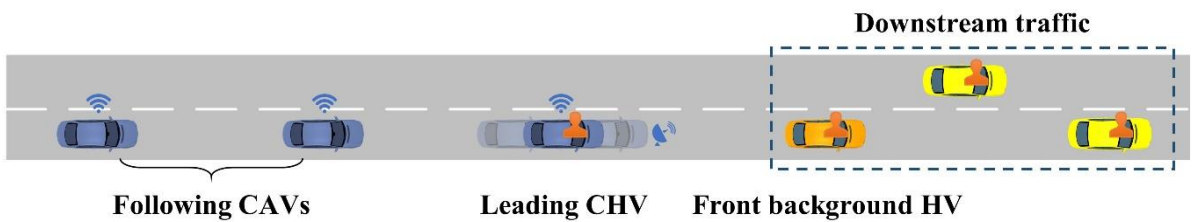

**Fig. 3.** The test scenario

To comprehensively evaluate the proposed SDHL-CACC controller, the following two cases are designed:

**Case 1**: Downstream traffic with fluctuating speed
**Case 2**: Downstream traffic with drastic speed reduction

Sensitivity parameters of the two cases are defined in **Table 1**.

**Table 1** Case definition

|  | **Downstream traffic states** | | |
|---|---|---|---|
|  | Acceleration/Deceleration (m/s$^2$) | Speed oscillation range(m/s) | Speed reduction range (m/s) |
| **Case 1** | ±1, ±2, ±3 | ±1, ±2, ±3 (around 17m/s) | / |
| **Case 2** | -4, -5, -6 | / | -3, -4, -5 |

*4.1.2 Controller Types*

- **The proposed SDHL-CACC controller**: The proposed controller is with the consideration of the leading CHV's uncertain behavior.
- **Baseline HL-CACC controller**: The baseline controller is a conventional MPC based HL-CACC controller (Zhang et al., 2022). This controller assumes that disturbances, the acceleration of the leading CHV, remain constant in the predictive horizon. It does not take into consideration the uncertainties from the leading CHV.

*4.1.3 Measurement of Effectiveness (MOE)*

The following MOEs are adopted.

- **Function validation**: The function of the controller is validated by *vehicle trajectories*, including location, velocity, and acceleration.
- **Perceived safety**: Perceived safety is mostly evaluated by traffic oscillations (Ha et al., 2020; Saifuzzaman et al., 2017; Wu and Wang, 2021). This paper analyzes traffic oscillation reduction performance from two aspects: the magnitude and the frequency. The magnitude of traffic oscillation reduction is quantified by the *acceleration range of the following CAVs* $R_a$. The frequency of traffic oscillation reduction is evaluated based on the *frequency-domain analysis* of the following CAVs' acceleration (Frigo and Johnson, 1998). Furthermore, this study quantifies the perceived safety based on Time-to-Collision (TTC) (Kiefer et al., 2005). The *perceived safety indicator* $P_s$ is designed as follows. A greater $P_s$ means higher perceived safety.

$$P_s = c \times \frac{1}{1 + \exp(-b + w \times TTC)} \quad (30)$$

where $TTC = \frac{x_1 - x_2}{-(v_1 - v_2)} = \frac{\Delta x}{\Delta v}$, $c = 1.0$, $b = -2.2$, $w = 1.0$. $P_s$ ranges from 0 to 1 as shown in **Fig. 4**.

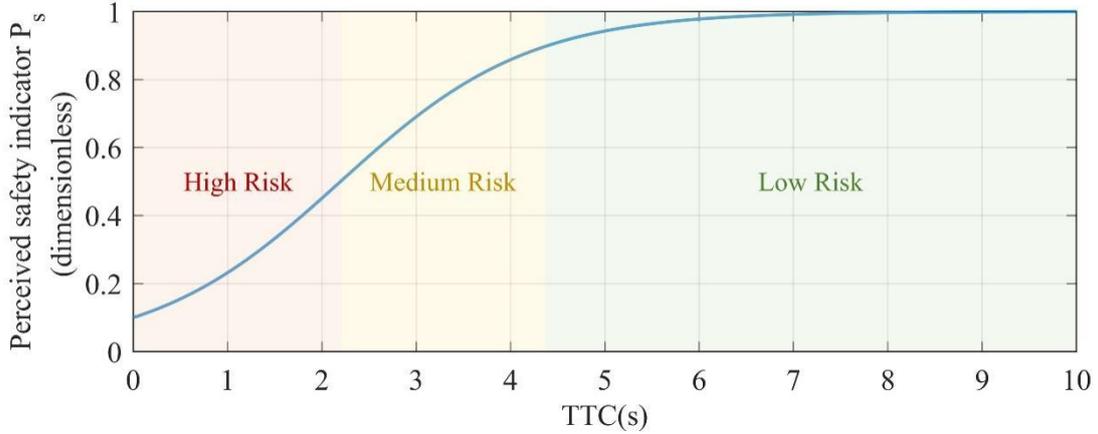

**Fig. 4.** Perceived safety indicator curve

- **Actual safety**: Actual risk is measured by *following distance* between adjacent vehicles.
- **String stability**: String stability is evaluated by the reduction of acceleration range along the platoon. *Oscillation transfer parameter* is defined as $O_t = \frac{acceleration\ range\ of\ follower2}{acceleration\ range\ of\ follower1}$. A smaller oscillation transfer value indicates better string stability.

*4.1.4 Parameter Settings*

The following settings are adopted for the evaluation. The proposed SDHL-CACC controller and the baseline HL-CACC controller share the same parameters (from #1 to #6), except the parameters in regard to uncertainty consideration (#7 and #8).

1) Desired platoon distance headway: 15m;
2) Acceleration range: $[-5,3]\ m/s^2$;
3) Lane width: 3.5m (11.5ft);
4) The maximum number of nodes in the optimization tree: $n_{max} = 50$ (control step: 0.1s, control horizon: 1s);
5) The weighting factors of state error: $[15,10,15,10,1]^T$;
6) The weighting factors of control cost: $r_u = 2$.
7) The accepted minimal distance error: $e = 2m$.
8) The weighting factor of the accepted minimal distance error: $M = 1000$.

In the evaluation, the behaving of the leading vehicle and the front background vehicle are calibrated according to real vehicle trajectory data as shown in **Fig. 5**. The real data is provided by the UTE team from Southeast University, containing the detailed vehicle traveling information collected on a China freeway. A standard genetic algorithm is used to calibrate the model parameters (Ngoduy et al., 2019). Calibrated parameters in the stochastic driver model (equation (11)) are shown in **Table 2**.

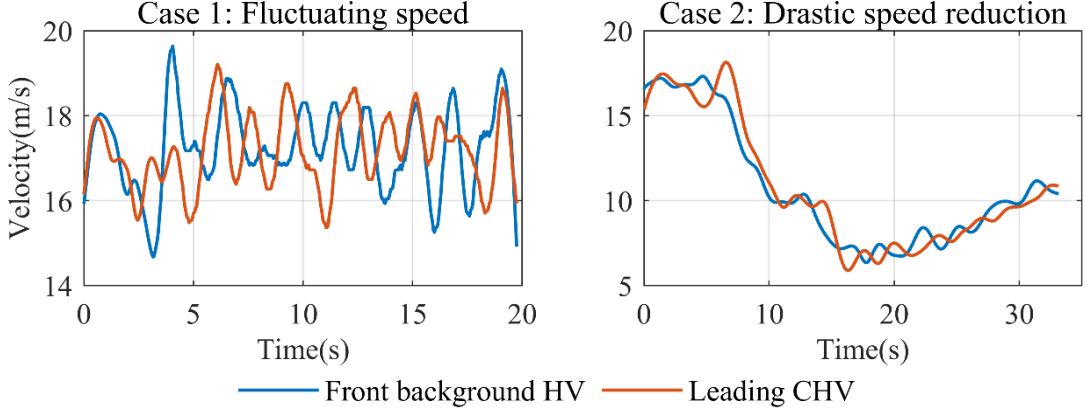

**Fig. 5.** Real vehicle trajectory data for two cases

**Table 2** The stochastic driver model parameters

| Parameters | Definition | Value |
|---|---|---|
| $v_0$ | Free flow speed $(m/s)$ | 19.65 |
| $\beta$ | Reaction coefficient $(1/s)$ | 1.92 |
| $s_c$ | Critical headway $(m)$ | 5.38 |
| $\alpha$ | Constant coefficient (dimensionless) | 2.66 |
| $\sigma_0$ | Dissipation coefficient $(\sqrt{m}/s)$ | 0.30 |

**4.2 Results**

Results show that the proposed SDHL-CACC controller can i) improve the perceived safety by 19.17% in traffic with oscillation; ii) enhance actual safety of the platoon by 7.76% against hard brakes; iii) guarantee string stability in all situations. The average computation time is approximately 3.2 milliseconds on a laptop equipped with an Intel i5-13500H CPU, indicating that the proposed controller is ready for real world implementation.

*4.2.1 Function Validation*

Vehicle trajectories of the proposed SDHL-CACC controller and the baseline HL-CACC controller are illustrated in **Fig. 6**. As shown in **Fig. 6** (a), (b), (g), and (h), the two controllers both ensure CAVs in a platoon maintain a stable following space. As shown in **Fig. 6** (c), (d), (e), and (f), when the leading CHV's speed fluctuates, the proposed SDHL-CACC controller is capable of stably maneuvering the following CAVs. However, the baseline HL-CACC controller maneuvers the following CAVs fluctuating with the leading CHV. Hence, the proposed SDHL-CACC controller is capable of relieving traffic fluctuations, thereby enhancing platooning safety. As shown in **Fig. 6** (i), (j), (k), and (l), in **Case 2**-drastic speed reduction, the leading CHV stochastically accelerates during the speed reduction process, highlighted in the green circles. The conventional HL-CACC quickly responds by accelerating to follow the leading CHV, as shown in **Fig. 6** (i). However, the leading CHV's acceleration is merely a stochastic behavior, which is soon followed by another drastic speed reduction. Our proposed controller could anticipate the leading CHV's decelerating motions and maneuver the followers to proactively slow down, as depicted in **Fig. 6** (j). Thus, the proposed SDHL-CACC controller is able to guarantee safety under hard deceleration situations.

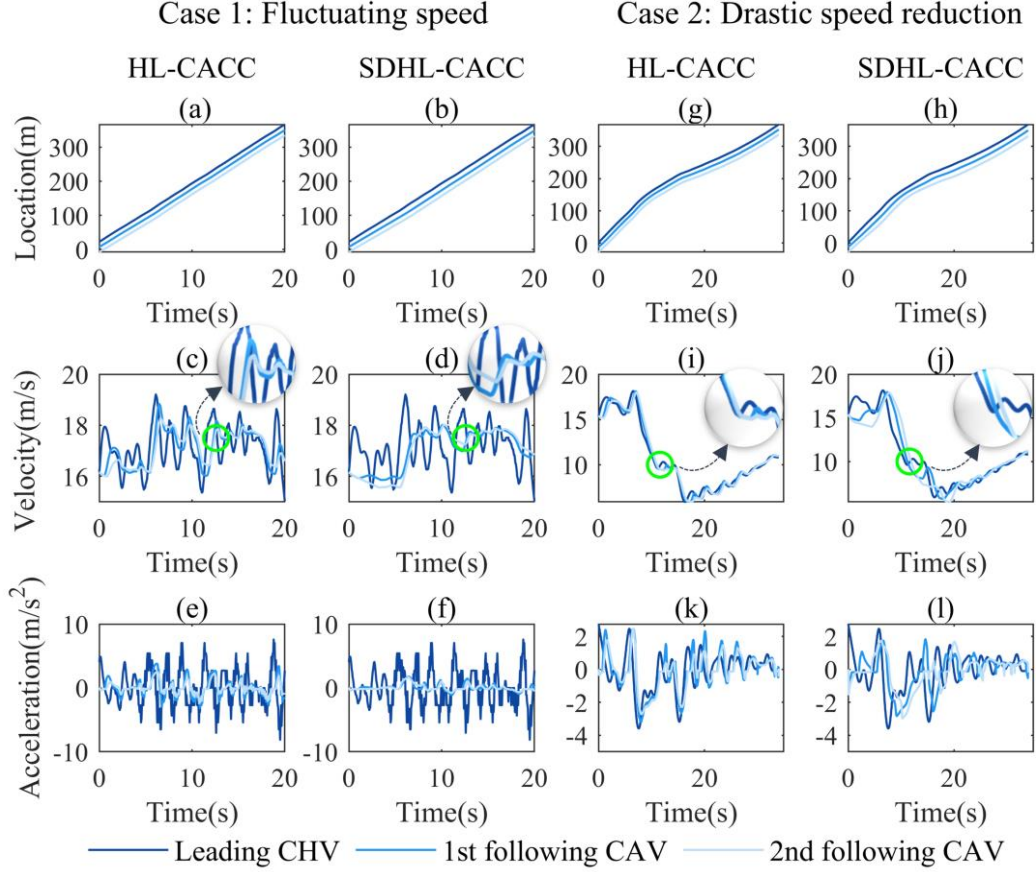

**Fig. 6.** Trajectories of the platoon in two cases

The rationale of the proposed controller's proactive reaction is justified by the uncertainty prediction results shown in **Fig. 7**. It illustrates the predicted acceleration at the position highlighted by the green circle in **Fig. 6** (c), (d), (i), and (j). In **Fig. 7**, each point represents a node in the optimization tree. The color of points is mapped with the magnitude of acceleration. Point size reflects the probability of the node. Results show that the baseline HL-CACC controller assumes that the current state of the leading CHV would continue. Conversely, the SDHL-CACC controller could consider all possible motions of the leading CHV. In **Case 1**-speed fluctuating, as shown in **Fig. 7** (a) and (b), the HL-CACC assumes that the leading CHV would continue its current deceleration behavior. Comparably, the proposed SDHL-CACC controller predicts the probability of the leading CHV's all possible motions, including both acceleration and deceleration. This enables the proposed SDHL-CACC controller to reduce speed oscillations in the platoon, as shown in **Fig. 6** (d). In **Case 2**-drastic speed reduction, the leading CHV may conduct an accelerating motion during the process of drastic speed reduction, as highlighted by the green circle in **Fig. 6** (i) and (j). At this time, the conventional HL-CACC assumes that the leading CHV would continue to accelerate, as shown in **Fig. 7** (c). It leads the followers to accelerate and follow up with the leading CHV, as shown in **Fig. 6** (i). Our proposed SDHL-CACC controller just solves this problem. As shown in **Fig. 7** (d), the proposed SDHL-CACC captures the leading CHV's decelerating motions which may be more likely to happen in the future. Hence, SDHL-CACC could maneuver the followers to proactively decelerate, instead of closely following up with the leading CHV's current accelerating motions, as shown in **Fig. 6** (j). The capability of capturing the leading CHV's uncertainties greatly enhances driving safety of CACC.

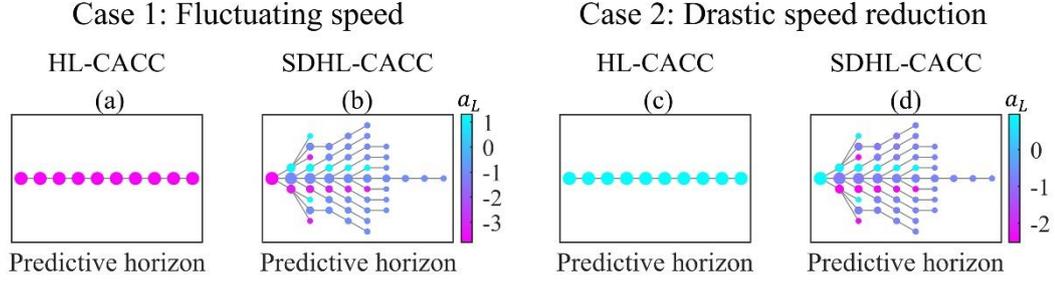

Fig. 7. Uncertainty prediction comparison in two cases

*4.2.2 Perceived Safety Quantification*

The SDHL-CACC controller is demonstrated to enhance perceived safety by reducing traffic oscillations. The magnitude of traffic oscillation is quantified by the acceleration range of the following CAVs, as illustrated in **Fig. 8**. It demonstrates that the proposed SDHL-CACC controller enhances perceived safety due to the reduction of the acceleration range by 58.82% for the first following CAV and 49.68% for the second following CAV, compared to the conventional HL-CACC controller. Furthermore, a sensitivity analysis is conducted for the magnitude of traffic oscillation in terms of acceleration and speed oscillation range of the front background HV, as shown in **Fig. 8** (b), Compared to the baseline HL-CACC controller, the proposed SDHL-CACC controller exhibits greater robustness against oscillations. Moreover, an interesting finding is that the acceleration range of the following CAVs increases with the decrease of acceleration range of the front background HV, and increases with the increase of speed oscillation range of the front background HV. It makes sense since a greater speed range and less acceleration range of the front background HV means larger oscillation waves (higher amplitude and less frequent vibration).

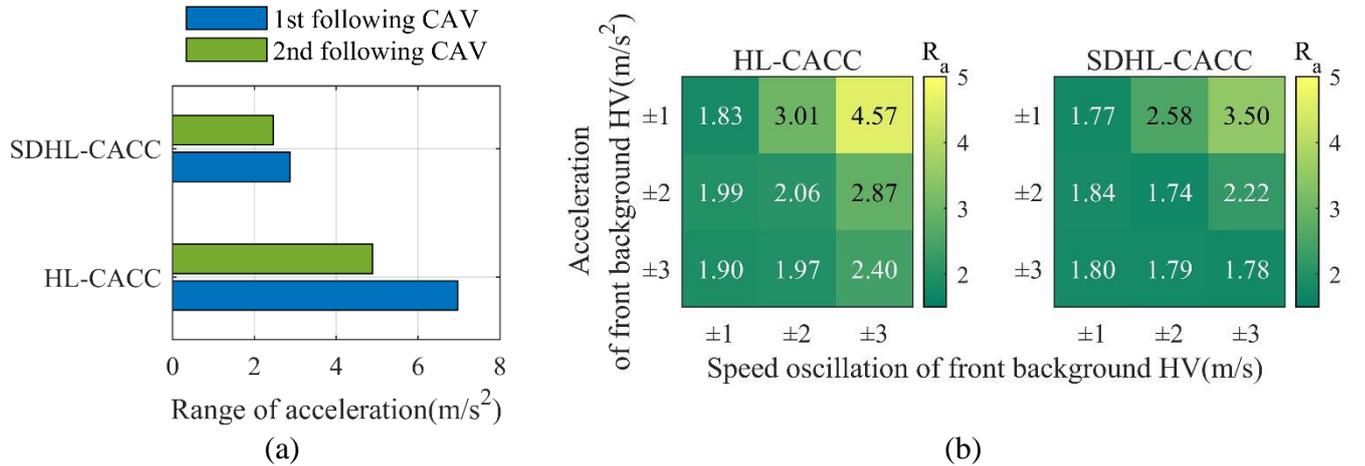

Fig. 8. Acceleration range of following CAVs

The frequency reduction of traffic oscillation is evaluated based on the frequency-domain analysis of the following CAVs, as illustrated in **Fig. 9**. The frequency-domain analysis is based on the discrete Fourier transform (DFT), which is efficiently computed using the Fast Fourier Transform (FFT) algorithm (Frigo and Johnson, 1998). It shows that the proposed SDHL-CACC controller enhances perceived safety by reducing the frequency of acceleration oscillations. Compared to the baseline HL-CACC controller, the proposed SDHL-CACC controller has more low-frequency oscillations and much less high-frequency oscillations. The reduction in traffic oscillation is quantified by the average magnitude and frequency for a clearer illustration, as shown in **Fig. 10**. The SDHL-CACC controller shows a significant improvement compared to the baseline HL-CACC controller, with reductions of 53.30% in average magnitude and 66.55% in average frequency. This phenomenon demonstrates that the SDHL-CACC controller substantially minimizes driving oscillations from the leading CHV, resulting in smoother and more stable cruising.

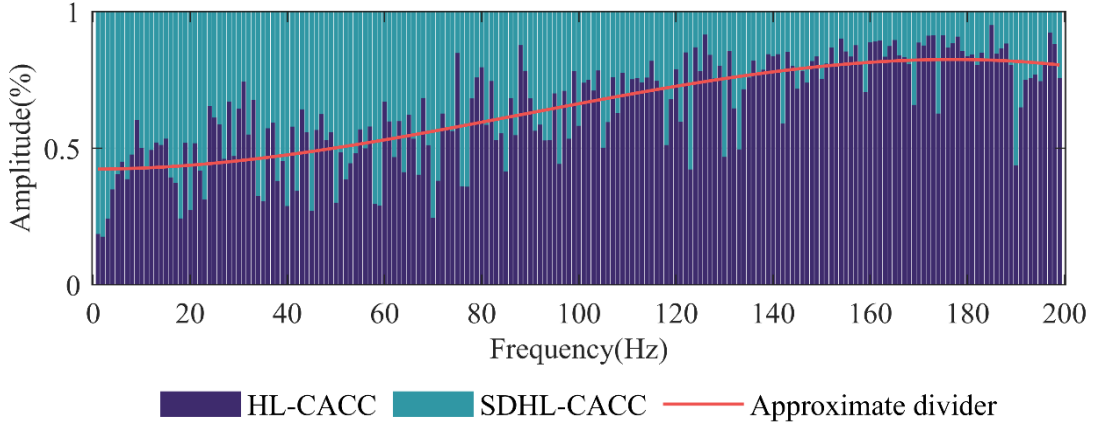

**Fig. 9.** Frequency analysis of the following CAVs' acceleration

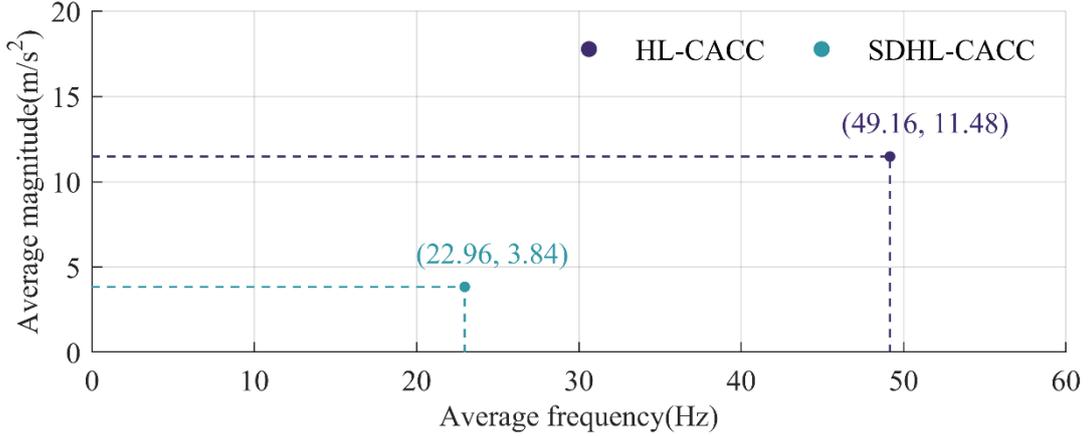

**Fig. 10.** Quantified oscillation of following CAVs' acceleration

The perceived safety is quantified by the perceived safety indicator $P_s$ defined in equation (30). **Fig. 11** illustrates the minimal perceived safety indicator $P_s$ for the two controllers. It is shown that the proposed SDHL-CACC controller enhances the minimal perceived safety indicator by 22.12% compared to the conventional HL-CACC controller. The proposed SDHL-CACC controller ensures the minimal perceived safety indicator is always greater than 0.5. According to **Fig. 4**, the proposed SDHL-CACC controller is capable of limiting risks within the medium domain. However, the baseline HL-CACC controller would suffer from high risks.

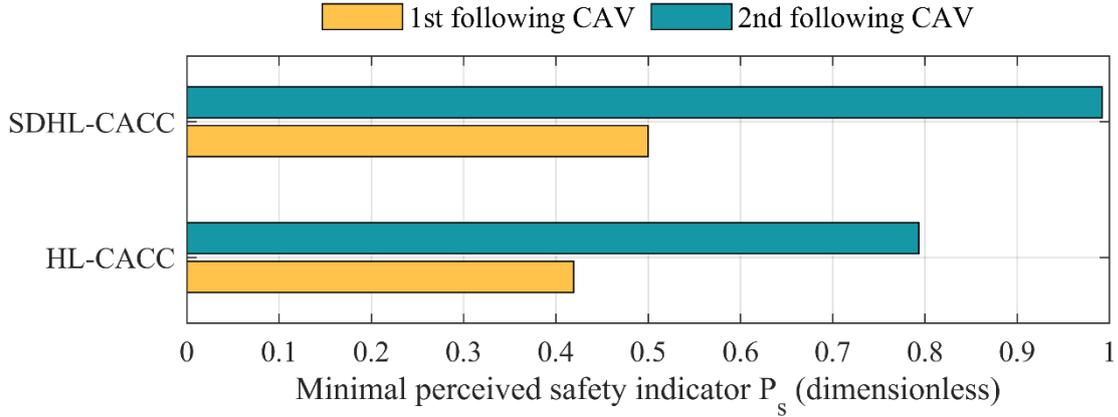

Fig. 11. Perceived safety quantification

*4.2.3 Actual Safety Quantification*

The proposed SDHL-CACC controller is confirmed with enhanced actual safety by maintaining a greater minimal following distance, as illustrated in **Fig. 12**. It demonstrates that the proposed SDHL-CACC controller enhances the minimal following distance by an average of 5.49%, compared to the baseline HL-CACC controller. Furthermore, a sensitivity analysis is conducted in terms of acceleration and speed range of the front background HV, as shown in **Fig. 12** (b). It reveals that the proposed SDHL-CACC controller is capable of maintaining the desired following distance (15 meters) when the speed reduction of the front background HV is less than 6m/s. Only a small decrease of the following distance (less than 1.22m/s) is found in great oscillations where the speed reduction of the front background HV is 6m/s.

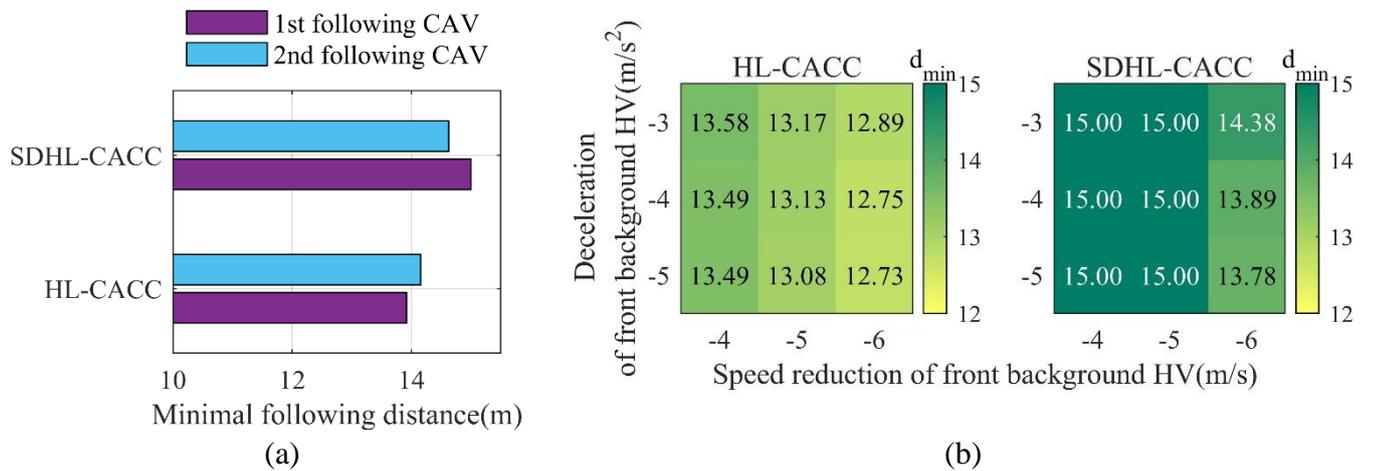

(a)                      (b)

Fig. 12. Minimal following distance analysis

*4.2.4 String Stability Validation*

The proposed SDHL-CACC controller is confirmed with string stability by ensuring the oscillation transfer parameter is always smaller than 1. A sensitivity analysis is conducted for the oscillation transfer parameter in terms of the acceleration and speed range of the front background HV, as shown in **Fig. 13**. It confirms that the proposed SDHL-CACC controller is capable of guaranteeing string stability in all cases, including traffic speed fluctuating and drastic speed reduction. Furthermore, an interesting finding is that more frequent oscillation would deteriorate the string stability. A greater oscillation transfer parameter is found when the oscillation is more frequent (smaller acceleration range and smaller speed range).

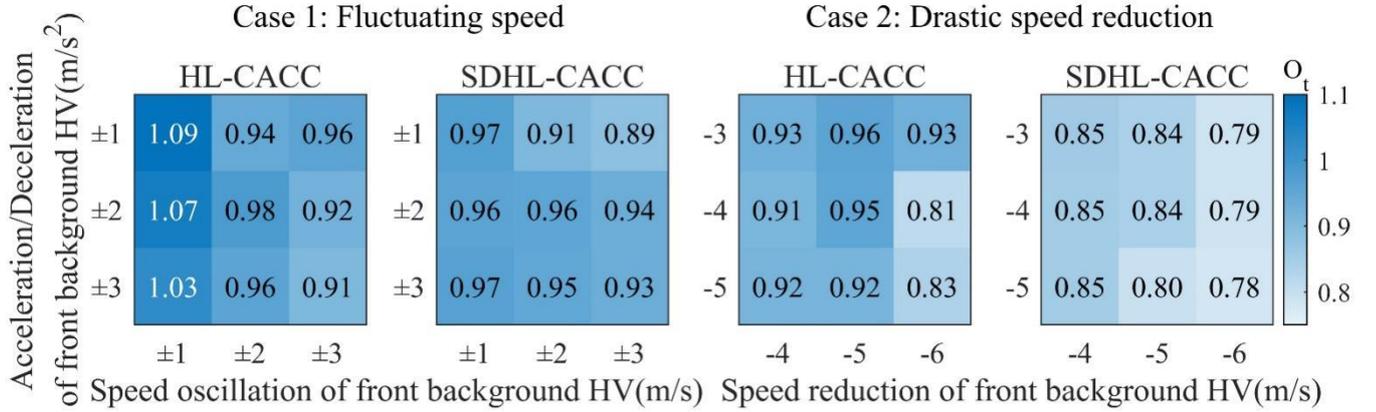

**Fig. 13.** Sensitivity analysis of oscillation transfer parameter

*4.2.5 Computation Efficiency Validation*

The proposed SDHL-CACC controller is confirmed to ensure real-time computation efficiency. A sensitivity analysis is conducted for computation efficiency in terms of the complexity of the optimization tree (measured by the number of nodes), as shown in **Fig. 14**. The proposed SDHL-CACC ensures an average computation time of 4.26 milliseconds on a laptop equipped with an Intel i5-13500H CPU. Increasing the complexity by 10 nodes enhances the computation time by 0.79 milliseconds. In our parameter setting, the proposed controller is functional under an optimization tree with 50 nodes, where the computation time is only 3.2 milliseconds. Even when complexity increases to 100 nodes, the computation can still be accomplished within 8.1 milliseconds. Hence, the real-time computational efficiency of the proposed controller could be guaranteed.

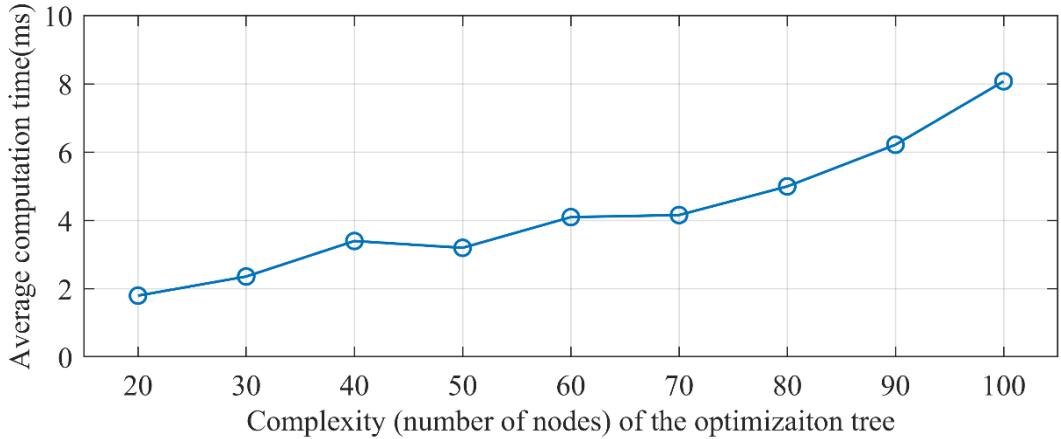

**Fig. 14.** Sensitivity analysis of computational efficiency

## 5 CONCLUSION AND FUTURE RESEARCH

In this paper, a HL-CACC controller is designed based on SMPC. It is enabled to predict the driving intention of the leading CHV. The proposed controller has the following features: i) enhanced perceived safety in oscillating traffic; ii) guaranteed safety against hard brakes; iii) computational efficiency for real-time implementation. The proposed controller is evaluated on a PreScan&Simulink simulation platform. Real vehicle trajectory data is collected for the calibration of the simulation. The result reveals that:
- The proposed controller functions as expected, effectively enabling several CAVs to maintain a consistent distance when following a human-driven leading vehicle.

- Compared to the conventional method, the proposed controller can improve perceived safety by 19.17% through avoiding higher oscillation frequency.
- Compared to the conventional method, the proposed controller enhances actual safety by 7.76%.
- The proposed controller is confirmed with string stability. Approximately 14.32% of oscillation is eliminated along the vehicle string in the platoon.
- The computation time of the proposed SDHL-CACC controller is approximately 3.2 milliseconds when running on a laptop equipped with an Intel i5-13500H CPU. This indicates the proposed controller has the potential for real-time implementation.

In this paper, the controller only considers the longitudinal disturbances of the leading CHV. Future studies could explore the impacts of background vehicles on the leading CHV's lateral driving behaviors. Besides, since the proposed stochastic driver model is calibrated offline using historical vehicle trajectory data, exploring methods to maintain the performance of the proposed SDHL-CACC controller under varying individual driving behavior is crucial. Future studies could consider enhancing this model by incorporating real-time trajectories of the lead CHV's driver. An online optimized prediction model shall effectively address driving behavior diversity. These enhancements would further improve the control accuracy and stability.

## 6 ACKNOWLEDGMENT

This paper is partially supported by National Key R&D Program of China (No. 2022YFF0604905), National Natural Science Foundation of China (No. 52302412 and 52372317), Yangtze River Delta Science and Technology Innovation Joint Force (No. 2023CSJGG0800), Shanghai Automotive Industry Science and Technology Development Foundation (No. 2404), Xiaomi Young Talents Program; the Fundamental Research Funds for the Central Universities, Tongji Zhongte Chair Professor Foundation (No. 000000375-2018082), the Postdoctoral Fellowship Program (Grade B) of China Postdoctoral Science Foundation (GZB20230519), Shanghai Sailing Program (No. 23YF1449600), Shanghai Post-doctoral Excellence Program (No.2022571), China Postdoctoral Science Foundation (No.2022M722405), and the Science Fund of State Key Laboratory of Advanced Design and Manufacturing Technology for Vehicle (No. 32215011).